\documentclass[conference]{IEEEtran}
\IEEEoverridecommandlockouts
\usepackage{cite}
\usepackage{amsmath,amssymb,amsfonts}
\usepackage{algorithmic}
\usepackage{graphicx}
\usepackage{textcomp}
 \usepackage{hyperref}
 \hypersetup{
    colorlinks=true,
    linkcolor=blue,
    filecolor=magenta,      
    urlcolor=blue,
}
\usepackage{adjustbox}
\usepackage{multirow}
\usepackage{verbatim}
\usepackage{pifont}
\newcommand{\cmark}{\ding{51}}%
\newcommand{\xmark}{\ding{55}}%

\usepackage{xcolor}
\def\BibTeX{{\rm B\kern-.05em{\sc i\kern-.025em b}\kern-.08em
    T\kern-.1667em\lower.7ex\hbox{E}\kern-.125emX}}
\begin{document}

\title{ More to Less (M2L): Enhanced Health Recognition in the Wild with Reduced Modality of Wearable Sensors}

\author{Huiyuan Yang$^{1}$, Han Yu$^{1}$, Kusha Sridhar$^{1}$, Thomas Vaessen$^{2}$, Inez Myin-Germeys$^{2}$ and Akane Sano$^{1}$
\thanks{*This work was supported by NSF \#1840167 and \#2047296}
\thanks{$^{1}$ H. Yang, H. Yu, K. Sridhar and A. Sano are with the Department of Electrical Computer Engineering, Rice University, Houston TX 77005, USA. Email: {\tt\small \{hy48, hy29, kh82,  Akane.Sano\}@rice.edu}}%
\thanks{$^{2}$T. Vaessen and I. Myin-Germeys are with the Department of Neurosciences, KU Leuven, Leuven, Belgium. Email: {\tt\small \{thomas.vaessen, inez.germeys\}@kuleuven.be}}%
}


\maketitle

\begin{abstract}

Accurately recognizing health-related conditions from wearable data is crucial for improved healthcare outcomes. To improve the recognition accuracy, various approaches have focused on how to effectively fuse information from multiple sensors.
Fusing multiple sensors is a common scenario in many applications, but may not always be feasible in real-world scenarios. 
For example, although combining bio-signals from multiple sensors (i.e., a chest pad sensor and a wrist wearable sensor) has been proved effective for improved performance, wearing multiple devices might be impractical in the free-living context.
To solve the challenges, we propose an effective more to less (M2L) learning framework to improve testing performance with reduced sensors through leveraging the complementary information of multiple modalities during training.
More specifically, different sensors may carry different but complementary information, and our model is designed to enforce collaborations among different modalities, where positive knowledge transfer is encouraged and negative knowledge transfer is suppressed, so that better representation is learned for individual modalities. 
Our experimental results show that our framework achieves comparable performance when compared with the full modalities. Our code and results will be available at \href{https://github.com/comp-well-org/More2Less.git}{https://github.com/comp-well-org/More2Less.git}.
\end{abstract}

\section{Introduction}
Wearable sensors are unobtrusive, affordable and user-friendly, making them suitable for continuous and ubiquitous monitoring of individual's physiological and behavioral profiles in the free-living context and providing valuable insights into individuals' health and fitness status for health and medical applications.
\begin{figure}[t]
     \centering
     \includegraphics[width=0.8\linewidth]{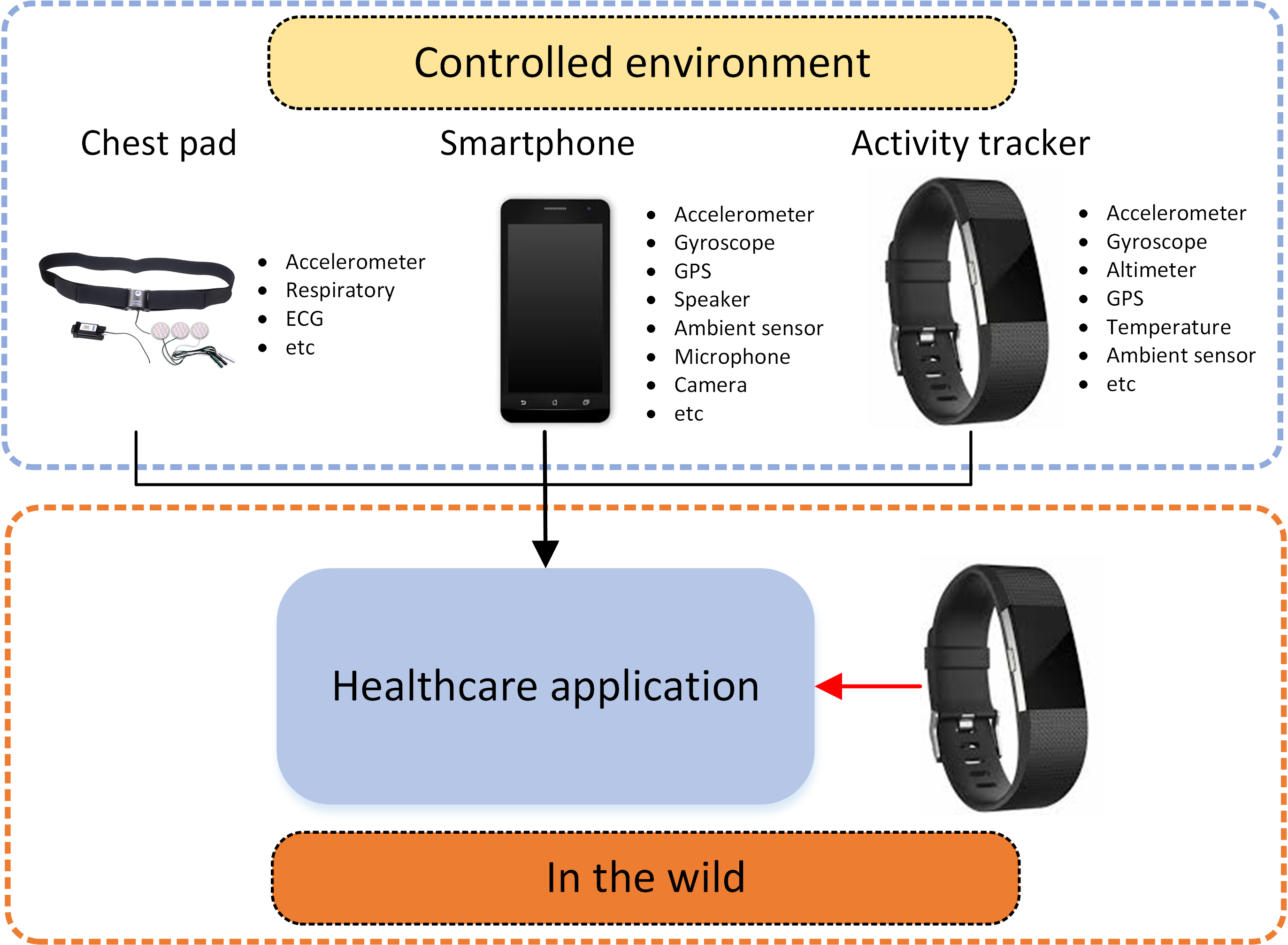}
       \caption{Illustration of the targeted challenge. In the controlled setting, we are usually able to collect various signals using multiple devices, such as standalone sensors, phone and fitbit, but less number of sensors or devices is more convenient for users in real-world settings. \textit{How can we design a framework that can leverage the benefits of multimodal data during training but maintain the robustness of model performance with reduced number of modalities during testing?}}
       \label{fig:motivation}
       \vspace{-0.5cm}
 \end{figure}
These advantages have attracted more and more researchers to adopt multimodal machine learning to  various wearable devices for better health monitoring and interventions,
as fusing data from different modalities can aggregate more information, therefore outperforming their unimodal counterparts. Huang et al. \cite{huang2021makes} provide appealing formal guarantees about the performance advantages of multimodal learning in comparison with unimodal learning using theoretical proofs. Some examples of multimodal learning frameworks proposed in the literature have fused audio and visual information for speech recognition \cite{shi2022learning}, improved word embeddings with both text and visual information\cite{mao2016training} and learned joint representations from text, visual and audio modality for sentiment analysis. Moving our attention to the field of wearable sensors, researchers have attempted to infer health related constructs or clinical events, from multimodal signals(\textit{i.e.,fitness trackers, smartwatches and smartphones}), for example, mental health and wellbeing monitoring \cite{yu2021modality,elgendi2019assessing, jaques2015multi}, seizure forecasting\cite{nasseri2021ambulatory}, COVID-19 detection \cite{quer2021wearable} and more applications;
exploiting the relationship among different modalities for better representation learning from  wearable data \cite{spathis2021self}; and
using additional sensors to improve single-sensor based complex activity recognition\cite{lago2021using}. 

Although research using multimodal modeling and wearable sensor technologies have increased with the advent of deep learning, only a small number of these studies have been successfully applied in our society.
Some challenges hinder the widespread adoption of wearable devices in healthcare applications, including data acquisition and pre-processing, feature extraction, and model selection. However, one specific challenge that we face in many real-world applications is the reduced availability of sensor modalities or devices in deployment compared to model training, and the challenge is understudied in the literature.

A common assumption for most works is that we have access to an equal number of sensors in both training and testing. 
However, such an assumption does not hold true in many circumstances in real-world scenarios.
For example, as shown in Fig.\ref{fig:motivation}, it is usually feasible to collect multiple modalities of data from study participants using different sensors (i.e., standalone sensors, chest sensors, wearables) in the controlled environment such as laboratory experiments. Therefore, we may be able to develop a robust multimodal model through effective fusion of complementary information from multiple modalities. 
However, in real deployment, using less number of sensors or devices is preferred, as we can therefore minimize user burden, energy consumption, or device size. (\textit{i.e., fitbit only}). 


Therefore, it is critical to bridge the gap between the models developed using multiple sensors during development and the models using less number of sensors during deployment in the wild.
In this work, we present an efficient framework, which not only leverages the complementary information of multiple modalities during training, but has the ability to provide inference with fewer modalities during testing (\textit{simulation of the model real-world deployment}). More specifically, an adaptive gate is designed for the multi-modalities, which will control the direction and intensity of knowledge transfer among modalities.  Therefore, positive knowledge transfer is encouraged, while negative knowledge transfer is suppressed. After training, we can thus expect improved performance for individual modality. 
Our main contributions are:
\begin{itemize}
    \item We propose an effective M2L framework that not only can leverage the complementary information of multiple modalities during training but also provide inference with fewer modalities during testing.
    \item We conduct extensive experiments using two wearable datasets, and the results demonstrate that our framework can benefit from the multimodal training, achieving comparable performance in testing with reduced modalities.
\end{itemize}

\vspace{-0.3cm}
\section{Propose Method}
We proposed an effective \textbf{M}ore to \textbf{L}ess (\textbf{M2L}) framework, which is designed to learn robust representations for each modality through leveraging the specific strengths existed in different modalities. This is accomplished by utilizing a cooperative learning strategy, where a weak network learns representations from a stronger network through knowledge distillation. More specifically, assuming $M$ modalities and $M$ classifier networks with similar architectures are available during training, with each classifier trained with its own data, it will also try to learn representations from other classifiers that show better performance than itself. The knowledge sharing is applied to the representations through minimizing the distance between the two features in the embedding space. In order to guarantee positive knowledge transferring, an adaptive regularizer is applied to ensure that knowledge only transfers from more accurate modality networks to those with less accuracy, and not the other way.
 
 \begin{figure}[t]
     \centering
     \includegraphics[width=0.8\linewidth]{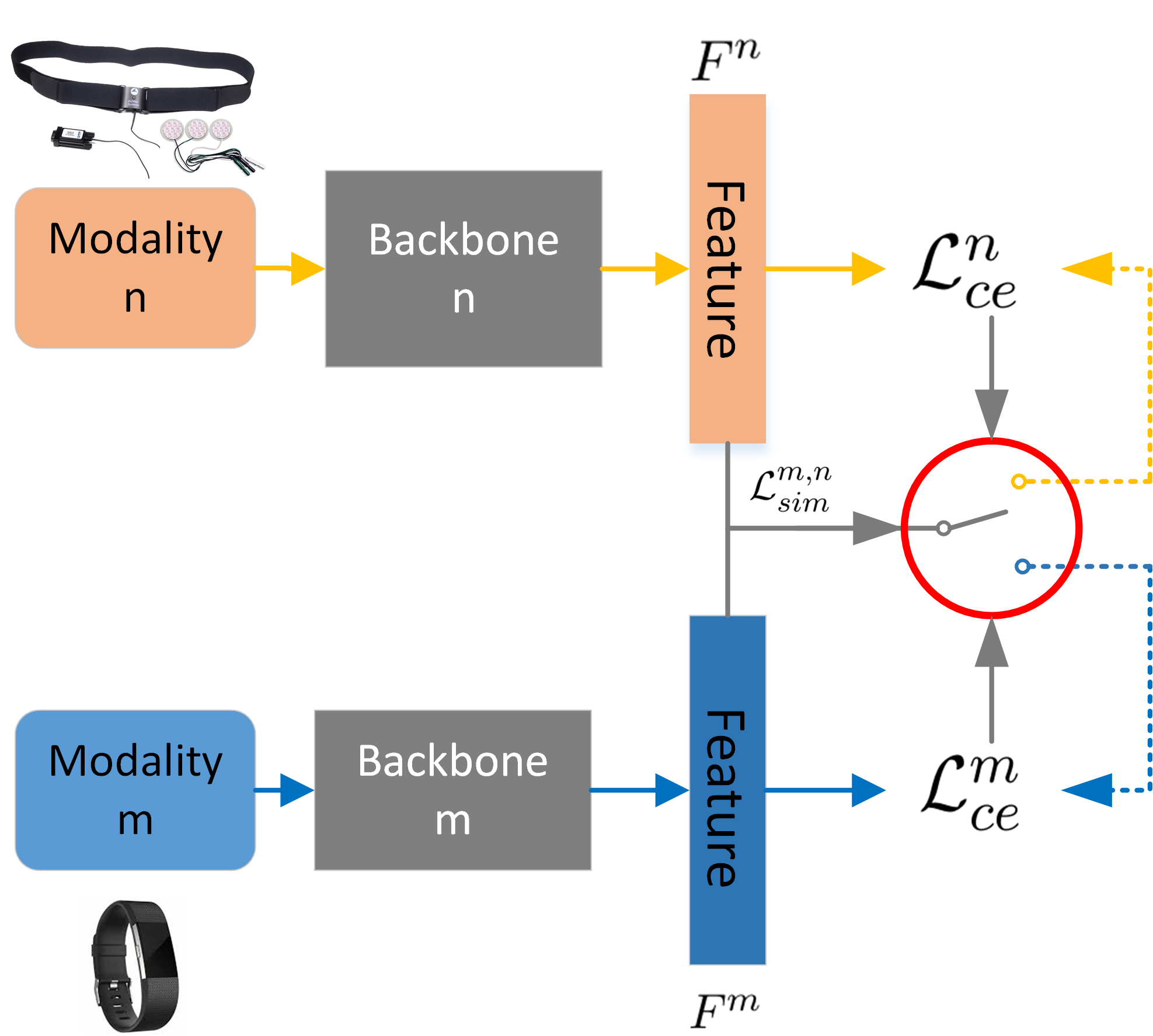}  
     \caption{Framework of the proposed M2L method.  Assuming we want to train backbone $m$ for the $m$th modality  with the information shared in $n$th modality. First, the cross-entropy classification loss, $\mathcal{L}_{ce}^{m}$ and $\mathcal{L}_{ce}^{n}$ are calculated for each modality. Next, $\rho ^{n \rightarrow m}$ and $\rho^{m \rightarrow n}$, determining the direction and intensity for distillation-based regularization, is calculated through comparing the two loss functions $\mathcal{L}_{ce}^{m}$ and  $\mathcal{L}_{ce}^{n}$.  The full objective function for training backbone $m$ for the $m$th modality is the weighted combination of $\mathcal{L}_{ce}^{m}$ and $\mathcal{L}_{sim}^{m, n}$.  After training, the proposed framework can be run with reduced modalities, while leveraging the benefits of multimodal training.}
       \label{fig:multimodal_training_unimodal_testing}
       \vspace{-0.3cm}
 \end{figure}
 
\subsection{Classification Loss and Feature Distance}
Let $\mathcal{D} = \{(x_{i}, y_{i})\}^{N}$ be a multimodal dataset having $N$ training samples. Each sample $x_{i}$ represents the data of $M$ available modalities, $x_{i} = \{ x_{i}^{1}, x_{i}^{2}, \dots x_{i}^{M} \}$, and $y_i$ represents its label. For each modality of $x_{i}^{m}$, a backbone network $\mathbf{f}^{m}(\cdot)$ is used to map the input into feature space: $\mathbf{f}^{m}: x_{i}^{m} \rightarrow F_{i}^{m}$, and $F_{i}^{m} \in \mathbb{R}^{d}$.
The supervised classification loss with respect to a specific modality and network (i.e. $m$th modality) is defined as:
\begin{equation}
 \mathcal{L}_{ce}^{m} = -\frac{1}{N} \sum_{i=1}^{N} \bigg\{ y_{i} \times \log (\hat{y}_{i}) + (1 - y_{i}) \times \log (1 - \hat{y}_{i})   \bigg\}
\end{equation}

 To benefit from the complementary information in the multiple modalities, we encourage the networks of the $m$th and the $n$th to share their own advantages 
 with each other. This can be done through minimizing their distance in the feature space, but experiments suggest that directly minimizing the L2 distance in the feature space will lead to unstable training. 
 As a result, we choose the cosine similarity as the metric:
 \begin{equation}
     \mathcal{L}_{sim}^{m,n} = \frac{1}{N} \sum_{i=1}^{N} \frac{F_{i}^{m}\cdot (F_{i}^{n})^{T}}{||F_{i}^{m}|| \cdot ||F_{i}^{n} || } \ \ \  m, n \in \{1, 2, \dots M\}
 \end{equation}
 
 \subsection{Transferring Positive Knowledge}
 As mentioned before, different modalities may convey varied information, and some modalities may provide weak features as compared to the others and vice versa.  In addition, even the strong modalities may sometimes have corrupted examples such as noise in the training dataset. With these cases in mind, it is desirable to develop a method that encourages positive knowledge transfer between the networks while avoiding negative transfer. Such a mechanism is implemented as $\rho (\cdot)$ in our framework. For example, as shown in Fig.\ref{fig:multimodal_training_unimodal_testing}, $\rho ^{n\rightarrow m}$ regulates the direction and intensity of transferring knowledge from modality $n$ to modality $m$. 

 Assume $\mathcal{L}_{ce}^{m}$ is the classification losses of the networks $m$. Next, let $\Delta \mathcal{L}^{i \rightarrow m} = \mathcal{L}_{ce}^{m} - \mathcal{L}_{ce}^{i}, i \in \{1, 2, \dots M\}$ be their difference. A positive $\Delta \mathcal{L}^{i \rightarrow m}$ indicates that network $i$ works better than network $m$. Hence, in the training of network $m$, we want $\rho^{i \rightarrow m}$ open the gate and transfer knowledge from network $i$ to network $m$, where the strength is conditioned on the value of $\Delta \mathcal{L}^{i \rightarrow m}$. On the other hand, a negative $\Delta \mathcal{L}^{i \rightarrow m}$ indicates that network $i$ is weaker than network $m$, so we want to avoid the knowledge transfer by setting $\rho^{i \rightarrow m} $ as $0$. The regularizer for training modality $m$ with the assistance of other modalities is defined as below:
 \begin{equation}
     \rho^{i \rightarrow m} = 
     \begin{cases}
            e^{\beta \Delta \mathcal{L}^{i \rightarrow m}} -1 &    \Delta \mathcal{L}^{i \rightarrow m}  > 0 \\
            0  &    \Delta \mathcal{L}^{i \rightarrow m} \leq 0  \ \ \  i \in \{1, 2,\dots M\}
     \end{cases}
 \end{equation}
 where $\beta$ is a positive hyper-parameter, which is used to control the strength of knowledge transferring.
 
 \subsection{Full Objective Function}
 Combining all the loss functions together, our full objective function for the training of network $m$ corresponding to the $m$th modality in a dataset containing $M$ modalities is defined as follows:
 \begin{equation}
     \mathcal{L}_{all}^{m} = \mathcal{L}_{ce}^{m} + \lambda \cdot\sum_{n = 1, n \neq m}^{M} \rho ^{n \rightarrow m} \cdot \big (1 - \mathcal{L}_{sim}^{m, n} \big )
 \end{equation}
 where $\lambda$ is a positive regularization parameter. Fig.\ref{fig:multimodal_training_unimodal_testing} shows an overview of how the features of $n $th modality assist the learning procedure of the $m$th modality.
 After training, the cross-modality knowledge transferring module is not needed any more, therefore the individual modality can be run independently.
 

 \section{Experiments}
We used two wearable multimodal datasets to evaluate our proposed framework.
 \vspace{-0.1cm}
 \subsection{Data}
\textbf{SMILE}\cite{smets2018towards}: 
Wearable sensors and self-report data collected from 41 healthy participants (36 females and 5 males) in a 10-day study. 
Two types of wearable sensors were used to collect both Galvanic Skin Response (GSR) (a \textit{wrist-worn} device, Chillband, IMEC, Belgium, sampling rate: 256 Hz) and electrocardiogram (ECG) (chest patch sensor, Health Patch, IMEC, Belgium, 256 Hz).
%
Both time and frequency domain statistical features related to human stress status\cite{kim2018stress, sharma2012objective} were extracted  every minute from GSR (12 features) and ECG data (8 features) (see more about features in \cite{smets2018towards, yu2021modality}). 
Self-reported stress levels (0 ("not at all") to 6 ("very")) were also collected 10 times daily as ecological momentary assessment that were spaced out roughly 90 minutes apart. 
We set stress levels greater than 1 as positive examples (55\%) and others as negative examples (45\%) in our experiments. We used prior 1 hour of GSR and ECG data (1 minute $\times$ 60 steps) to infer upcoming stress labels.

\textbf{TILES}\cite{mundnich2020tiles}: wearable, smartphone, and survey data collected from over 200 hospital workers (31.1\% of the participants were male and 68.9\% were female, and age: 21 - 65 years old). We used heart rate and step count data collected with the Fitbit Charge 2 (sampled every 1 minute) and ECG data collected with the OMSignal smart garment (15-second long ECG signal in 250 Hz every 5 minutes). We extracted 25 time and frequency-domain ECG features (see more in \cite{hrvanalysis}) and resampled Fitbit data every 5 minutes to align with the ECG features. Self-reported stress levels were annotated by participants in a 5-point scale, which is further binarized via a similar procedure as used in \cite{gaballah2021context} using the average z-score of individual's stress levels. We used 2 hours of the data (5 minute $\times$ 24 steps) to infer upcoming stress labels.

For both datasets, we randomly split  70\% of the participants of the data as a training set, and the rest as a test set to conduct subject-independent experiments, where data collected from individual subjects can only appear in either training or testing, but not both.

\begin{table}[htp!]
\centering
\caption{Accuracy and consistency ratio in two datasets. }
\label{table:experiment_multimodal_training_corr}
\adjustbox{max width=\textwidth}{
\begin{tabular}{l|l|l|l}
\hline
Dataset & Modality & Acc   & Ratio      \\ \hline \hline
\multirow{2}{*}{SMILE}   & GSR     & 53.9  & \multirow{2}{*}{59.0}   \\  \cline{2-3}
        & ECG     & 50.8   &  \\ \hline
\multirow{2}{*}{TILES}   & ECG    & 55.2   &  \multirow{2}{*}{59.2}   \\  \cline{2-3}
        & fitbit    & 57.5   &  \\ \hline
\end{tabular}
}
\vspace{-0.5cm}
\end{table}

\subsection{Implementation Details}
To model the long sequential data, we use long short-term memory (LSTM) as backbone. Each of the LSTM models is a two layers of LSTM with 64 as the number of features in the hidden state.
Drop out rate is set as 0.5 during training, and 0 during testing.
For the hypterparameters, $\lambda$ is set to 0.05, and $\beta = 2$.

The networks are trained from scratch for all the experiments, using Adam optimizer with an initial learning rate of 0.001. The learning rate is decayed after every 10 epochs by 0.1. Batch size is set 100, and the model is trained for 50 epochs with early stopping.
We start with pretraining individual modalities for 20 epochs, and then continue training with the knowledge transfer loss. We implement the model with the Pytorch framework and perform training and testing on the NVIDIA GeForce 3090 GPU. 

\begin{table}[htbp!]
\centering
\caption{F1 scores and accuracy of the proposed method are reported on the SMILE dataset for stress detection in the wild.}
\label{table:experiment_multimodal_training_SMILE}
\adjustbox{max width=\textwidth}{
\begin{tabular}{l|l|l|l|l|l}
\hline
Method & \begin{tabular}[c]{@{}l@{}}Training\\modality\end{tabular} & \begin{tabular}[c]{@{}l@{}}Testing\\modality\end{tabular} & \begin{tabular}[c]{@{}l@{}}Reduced\\ modality\end{tabular}& Acc   & F1      \\ \hline \hline
\multirow{2}{*}{LSTM} & GSR     & same & \xmark    & 53.9 & 64.1   \\  \cline{2-6}
  & ECG   & same  & \xmark   & 50.8 & 61.5   \\ \hline
LSTM (fusion) & GSR+ECG & same & \xmark     & 52.6 & 58.8   \\ \hline
\multirow{2}{*}{\textbf{M2L}}  & GSR + ECG & GSR  & \cmark   & 56.1 & 66.8 \\ \cline{2-6}
   & GSR + ECG  & ECG  & \cmark    & 53.1 & 64.5  \\ \hline
\end{tabular}
}
\vspace{-0.5cm}
\end{table}

\subsection{Results}
\begin{table}[htbp!]
\centering
\caption{F1 scores and accuracy of the proposed method in the TILES dataset for stress detection in the wild.}
\label{table:experiment_multimodal_TILES}
\adjustbox{max width=\textwidth}{
\begin{tabular}{l|l|l|l|l|l}
\hline
Method & \begin{tabular}[c]{@{}l@{}}Training\\modality\end{tabular}  & \begin{tabular}[c]{@{}l@{}}Testing\\modality\end{tabular} & \begin{tabular}[c]{@{}l@{}}Reduced\\ modality\end{tabular} & Acc   & F1     \\ \hline \hline
 \multirow{2}{*}{LSTM} & ECG     & same &  \xmark   & 55.7 & 65.8   \\  \cline{2-6} 
 & fitbit   & same   & \xmark   & 57.4 & 66.7   \\ \hline
LSTM (fusion) & ECG+fitbit& same  & \xmark & 54.6 & 61.5  \\ \hline
Paper \cite{tiwari2019breathing}  & BR & same & \xmark & 59.5 & 56.9 \\ \hline

Paper \cite{tiwari2019stress} & HRV & same & \xmark & 60.6 & 58.2 \\ \hline
\multirow{2}{*}{\textbf{M2L}} & ECG+fitbit  & ECG & \cmark     & 57.5 & 72.2  \\  \cline{2-6} 
  & ECG+fitbit & fitbit   & \cmark   &  58.6 & 70.3 \\ \hline
\end{tabular}}
\vspace{-0.5cm}
\end{table}

Various experiments are conducted to evaluate the performance of the proposed method.
Both accuracy and F1-score are reported in our experiments for performance evaluation.
First, we verify our claim that different modalities convey varied information. As shown in Table.\ref{table:experiment_multimodal_training_corr}, the accuracy changes over different modalities \textit{i.e., GSR, ECG and fitbit}. We also calculate the consistency ratio of individual modality based prediction, which is calculated by dividing the total number of testing examples by consistent prediction of two individual modality.
The consistency ratio is only around 60\% in both datasets, indicating varied information among different modalities.

The performance evaluation on the SMILE dataset is reported in Table.\ref{table:experiment_multimodal_training_SMILE}. GSR-based model achieves 3\% higher accuracy than ECG-based model, while the early fusion of GSR$+$ECG does not always perform better than the individual modality, indicating that even though the multiple modalities contain rich and complementary information, the benefits will not be achieved without a carefully designed fusion mechanism.
Compared to the models trained and tested with a single modality (\textit{GSR or ECG}), our model can be trained with multiple modalities (\textit{GSR and ECG}) and tested with reduced modality (\textit{GSR only or ECG only}), showing significant improvement (2.2\% and 2.3\% higher in accuracy respectively). The improved performance for reduced modality testing demonstrates the effectiveness of our proposed method.

Experiments conducted on TILES dataset are shown in Table.\ref{table:experiment_multimodal_TILES}, and we observe 1.8\% and 1.2\% improvement in accuracy when compare our model with LSTM on the ECG and fitbit modality respectively.
Comparing with recent stress detection works \cite{tiwari2019breathing, tiwari2019stress}, which were trained and tested in TILES datasets using breathing rate (BR) and heart rate variability (HRV) respectively collected with OMSignal Garment  (details about training and testing and label settings are not described so we are not able to reproduce their experiments),
our model not only achieves comparable performance, but also be able to run with reduced modality (\textit{a cheaper solution, i.e., fitbit}), while all other methods need exactly the same modalities between training and testing.

\section{Conclusion}
The reduced number of wearable sensors/devices during deployment/testing in the wild environment compared to model training is a challenge that hinders the adoption of wearable devices for healthcare.
To solve this problem, we present an effective M2L framework that can not only leverage the complementary information of multiple modalities during training, but also provide inference with reduced modalities during testing, while achieving comparable performance when compared with full modalities. Therefore, bridging the gap between model development and model deployment in the wild. It is worth noting that the proposed framework works for three or more modalities as well.

Although the improved performance, our proposed method is still limited by the use of hand-crafted features and the ability to generalize to new subjects. 
In the future, we plan to investigate both deep learning based feature learning  and  unsupervised personalization, where a general model can be automatically adapted to individual through utilizing a small number of unlabeled data.

{\small
\bibliographystyle{IEEEtran}
\bibliography{egbib}

\begin{thebibliography}{10}
\providecommand{\url}[1]{#1}
\csname url@samestyle\endcsname
\providecommand{\newblock}{\relax}
\providecommand{\bibinfo}[2]{#2}
\providecommand{\BIBentrySTDinterwordspacing}{\spaceskip=0pt\relax}
\providecommand{\BIBentryALTinterwordstretchfactor}{4}
\providecommand{\BIBentryALTinterwordspacing}{\spaceskip=\fontdimen2\font plus
\BIBentryALTinterwordstretchfactor\fontdimen3\font minus
  \fontdimen4\font\relax}
\providecommand{\BIBforeignlanguage}[2]{{%
\expandafter\ifx\csname l@#1\endcsname\relax
\typeout{** WARNING: IEEEtran.bst: No hyphenation pattern has been}%
\typeout{** loaded for the language `#1'. Using the pattern for}%
\typeout{** the default language instead.}%
\else
\language=\csname l@#1\endcsname
\fi
#2}}
\providecommand{\BIBdecl}{\relax}
\BIBdecl

\bibitem{huang2021makes}
Y.~Huang, C.~Du, Z.~Xue, X.~Chen, H.~Zhao, and L.~Huang, ``What makes
  multimodal learning better than single (provably),'' \emph{arXiv preprint
  arXiv:2106.04538}, 2021.

\bibitem{shi2022learning}
B.~Shi, W.-N. Hsu, K.~Lakhotia, and A.~Mohamed, ``Learning audio-visual speech
  representation by masked multimodal cluster prediction,'' \emph{arXiv
  preprint arXiv:2201.02184}, 2022.

\bibitem{mao2016training}
J.~Mao, J.~Xu, Y.~Jing, and A.~Yuille, ``Training and evaluating multimodal
  word embeddings with large-scale web annotated images,'' \emph{arXiv preprint
  arXiv:1611.08321}, 2016.

\bibitem{yu2021modality}
H.~Yu, T.~Vaessen, I.~Myin-Germeys, and A.~Sano, ``Modality fusion network and
  personalized attention in momentary stress detection in the wild,'' in
  \emph{9th Int. Conf. on Affective Computing and Intelligent Interaction
  (ACII)}.\hskip 1em plus 0.5em minus 0.4em\relax IEEE, 2021, pp. 1--8.

\bibitem{elgendi2019assessing}
M.~Elgendi and C.~Menon, ``Assessing anxiety disorders using wearable devices:
  Challenges and future directions,'' \emph{Brain sciences}, vol.~9, no.~3,
  p.~50, 2019.

\bibitem{jaques2015multi}
N.~Jaques, S.~Taylor, A.~Sano, and R.~Picard, ``Multi-task, multi-kernel
  learning for estimating individual wellbeing,'' in \emph{Proc. NIPS Workshop
  on Multimodal Machine Learning, Montreal, Quebec}, vol. 898, 2015, p.~3.

\bibitem{nasseri2021ambulatory}
M.~Nasseri, T.~Pal~Attia, B.~Joseph, N.~M. Gregg, E.~S. Nurse, P.~F. Viana,
  G.~Worrell, M.~D{\"u}mpelmann, M.~P. Richardson, D.~R. Freestone
  \emph{et~al.}, ``Ambulatory seizure forecasting with a wrist-worn device
  using long-short term memory deep learning,'' \emph{Scientific reports},
  vol.~11, no.~1, pp. 1--9, 2021.

\bibitem{quer2021wearable}
G.~Quer, J.~M. Radin, M.~Gadaleta, K.~Baca-Motes, L.~Ariniello, E.~Ramos,
  V.~Kheterpal, E.~J. Topol, and S.~R. Steinhubl, ``Wearable sensor data and
  self-reported symptoms for covid-19 detection,'' \emph{Nature Medicine},
  vol.~27, no.~1, pp. 73--77, 2021.

\bibitem{spathis2021self}
D.~Spathis, I.~Perez-Pozuelo, S.~Brage, N.~J. Wareham, and C.~Mascolo,
  ``Self-supervised transfer learning of physiological representations from
  free-living wearable data,'' in \emph{Proceedings of the Conference on
  Health, Inference, and Learning}, 2021, pp. 69--78.

\bibitem{lago2021using}
P.~Lago, M.~Matsuki, K.~Adachi, and S.~Inoue, ``Using additional training
  sensors to improve single-sensor complex activity recognition,'' in
  \emph{Int. Symposium on Wearable Computers}, 2021, pp. 18--22.

\bibitem{smets2018towards}
E.~Smets, ``Towards large-scale physiological stress detection in an ambulant
  environment,'' 2018.

\bibitem{kim2018stress}
H.-G. Kim, E.-J. Cheon, D.-S. Bai, Y.~H. Lee, and B.-H. Koo, ``Stress and heart
  rate variability: a meta-analysis and review of the literature,''
  \emph{Psychiatry investigation}, vol.~15, no.~3, p. 235, 2018.

\bibitem{sharma2012objective}
N.~Sharma and T.~Gedeon, ``Objective measures, sensors and computational
  techniques for stress recognition and classification: A survey,''
  \emph{Computer methods and programs in biomedicine}, vol. 108, no.~3, pp.
  1287--1301, 2012.

\bibitem{mundnich2020tiles}
K.~Mundnich, B.~M. Booth, M.~l’Hommedieu, T.~Feng, B.~Girault,
  J.~L’hommedieu, M.~Wildman, S.~Skaaden, A.~Nadarajan, J.~L. Villatte
  \emph{et~al.}, ``Tiles-2018, a longitudinal physiologic and behavioral data
  set of hospital workers,'' \emph{Scientific Data}, vol.~7, no.~1, pp. 1--26,
  2020.

\bibitem{hrvanalysis}
R.~Champseix, ``Heart rate variability analysis,''
  \url{https://github.com/Aura-healthcare/hrv-analysis}, 2018.

\bibitem{gaballah2021context}
A.~Gaballah, A.~Tiwari, S.~Narayanan, and T.~H. Falk, ``Context-aware speech
  stress detection in hospital workers using bi-lstm classifiers,'' in
  \emph{IEEE Int. Conf. on Acoustics, Speech and Signal Processing
  (ICASSP)}.\hskip 1em plus 0.5em minus 0.4em\relax IEEE, 2021, pp. 8348--8352.

\bibitem{tiwari2019breathing}
A.~Tiwari, S.~Narayanan, T.~H. Falk, and x.~x, ``Breathing rate complexity
  features for “in-the-wild” stress and anxiety measurement,'' in
  \emph{2019 27th European Signal Processing Conference (EUSIPCO)}.\hskip 1em
  plus 0.5em minus 0.4em\relax IEEE, 2019, pp. 1--5.

\bibitem{tiwari2019stress}
A.~Tiwari, S.~Narayanan, and T.~H. Falk, ``Stress and anxiety measurement"
  in-the-wild" using quality-aware multi-scale hrv features,'' in \emph{IEEE
  Engineering in Medicine and Biology Society (EMBC)}.\hskip 1em plus 0.5em
  minus 0.4em\relax IEEE, 2019, pp. 7056--7059.

\end{thebibliography}
}

\end{document}